\documentclass[twocolumn,hf]{ceurart}

\usepackage{listings}
\usepackage{comment}
\usepackage[scaled]{beramono}
\usepackage{microtype}
\usepackage{inconsolata}
\usepackage{cleveref}
\usepackage{graphicx}
\usepackage{array, multirow, bigdelim, makecell, booktabs} 
\usepackage{tablefootnote}

\begin{document}

\copyrightyear{2024}
\copyrightclause{Copyright for this paper by its authors.
  Use permitted under Creative Commons License Attribution 4.0
  International (CC BY 4.0).}

\conference{CLiC-it 2024: Tenth Italian Conference on Computational Linguistics, Dec 04 — 06, 2024, Pisa, Italy}

\title{Non Verbis, Sed Rebus: \\ Large Language Models are Weak Solvers of Italian Rebuses}

\author[1]{Gabriele Sarti}[%
    orcid=0000-0001-8715-2987,
    email=g.sarti@rug.nl,
    url=https://gsarti.com,
]
\cormark[1]

\author[1]{Tommaso Caselli}[%
orcid=0000-0003-2936-0256,
email=t.caselli@rug.nl,
]

\author[1]{Malvina Nissim}[%
orcid=0000-0001-5289-0971,
email=m.nissim@rug.nl,
]

\author[1]{Arianna Bisazza}[%
orcid=0000-0003-1270-3048,
email=a.bisazza@rug.nl,
url=https://cs.rug.nl/~bisazza
]

\cortext[1]{Corresponding author.}

\address[1]{Center for Language and Cognition (CLCG), University of Groningen, Oude Kijk in 't Jatstraat 26 \\\hspace{5pt} Groningen, 9712EK, The Netherlands}

\begin{abstract}
  Rebuses are puzzles requiring constrained multi-step reasoning to identify a hidden phrase from a set of images and letters. In this work, we introduce a large collection of verbalized rebuses for the Italian language and use it to assess the rebus-solving capabilities of state-of-the-art large language models. While general-purpose systems such as LLaMA-3 and GPT-4o perform poorly on this task, ad-hoc fine-tuning seems to improve models' performance. However, we find that performance gains from training are largely motivated by memorization. Our results suggest that rebus solving remains a challenging test bed to evaluate large language models' linguistic proficiency and sequential instruction-following skills.
\end{abstract}


\maketitle

\begin{figure}
    \centering
    \includegraphics[width=0.97\linewidth]{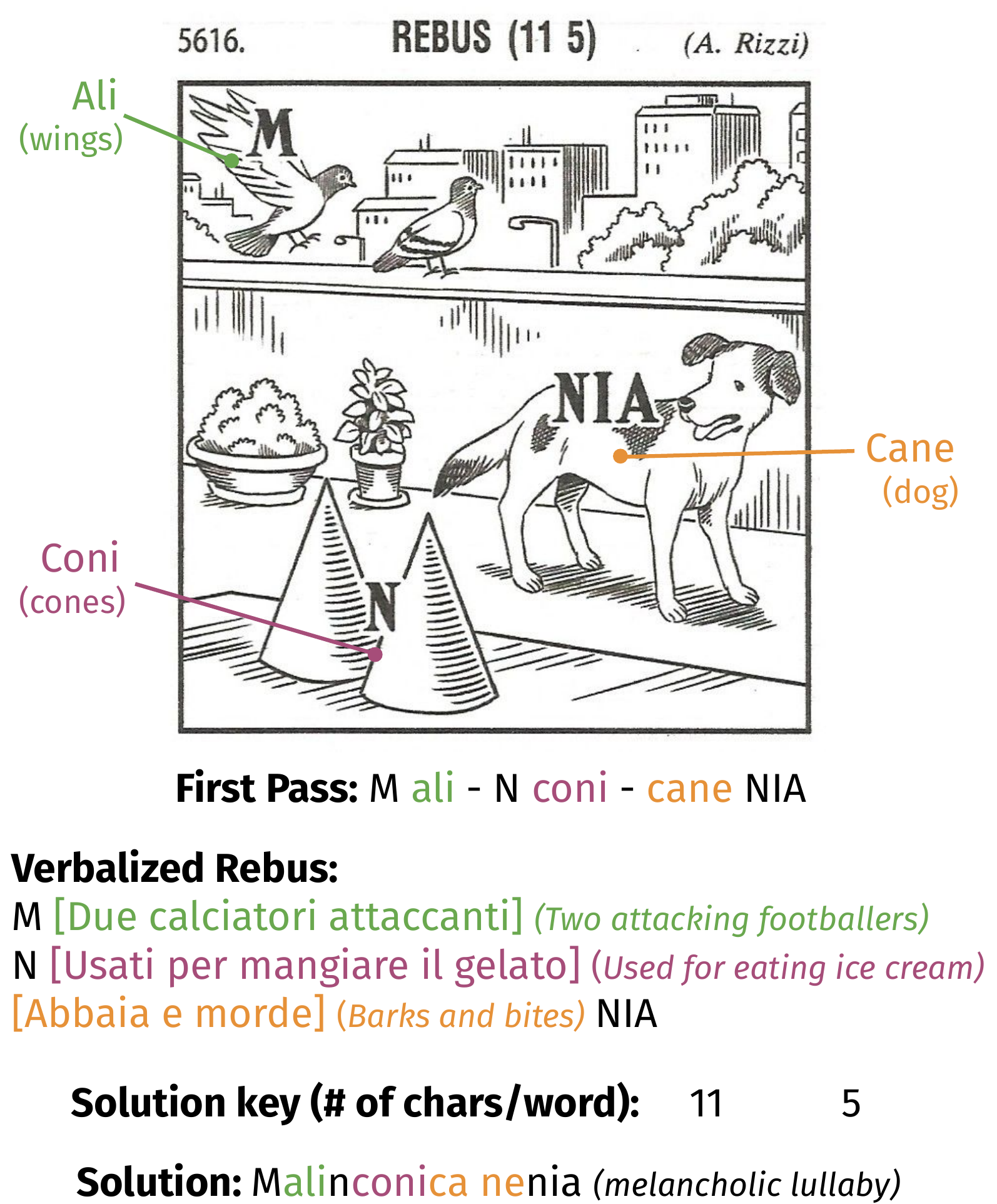}
    \caption{An example of a verbalized rebus crafted by combining a rebus first pass (intermediate solution) with crossword definitions. We use verbalized rebuses to test LLMs' sequential instruction following capabilities. Image from \textit{Settimana Enigmistica n. 4656}, \textcopyright~Bresi S.r.l.}
    \label{fig:example}
    \vspace{-12pt}
\end{figure}

\section{Introduction}

Complex games such as chess and Go have long been a source of inspiration to develop more flexible and robust AI systems~\citep{alphago,chess-ai}. Recent developments in NLP suggested that creative language games could be exploited as promising benchmarks for quantifying the ability of large language models (LLMs) to carry out multi-step knowledge-intensive reasoning tasks under pre-specified constraints~\citep{cryptic-crosswords}. While crossword puzzles have been historically the main focus of such efforts~\citep{english-crossword}, other categories of linguistic games received only marginal attention, especially for languages other than English. A prominent example of less-studied language games is the \textbf{rebus}, a visual puzzle combining images and graphic signs to encode a hidden phrase. Indeed, rebus solving is a complex, multi-step process requiring factual knowledge, contextual understanding, vocabulary usage, and reasoning within pre-defined constraints -- a set of fundamental skills to address a variety of real-world tasks.

In this work, we conduct the first open evaluation of LLMs' rebus-solving capabilities, focusing specifically on the Italian language. We propose a novel strategy to derive text-only \textit{verbalized rebuses} from transcribed intermediate rebus solutions and use it to produce a large collection with more than 80k verbalized rebuses. We then evaluate the rebus-solving skills of state-of-the-art LLMs, including open-source systems and proprietary models, via few-shot prompting. Moreover, we fine-tune a small but capable LLM on verbalized rebus solving, outperforming state-of-the-art systems by a wide margin. Finally, we conduct a fine-grained assessment of LLMs' sequential reasoning steps, explaining model performance in terms of word complexity and memorization.

Beyond rebus solving, our evaluation sheds light on the limits of current LLMs in multi-step reasoning settings, highlighting challenges with their application to complex sequential instruction-following scenarios.\footnote{Code, data and models are available on \href{https://github.com/gsarti/verbalized-rebus}{Github} and \href{https://huggingface.co/collections/gsarti/verbalized-rebus-clic-it-2024-66ab8f11cb04e68bdf4fb028}{Huggingface}}

%\AB{maybe motivate use of image descriptions instead of images}

%\MN{among contribs or in any case at end of intro I would also add a line on something like "what do we learn from this little exercise beyond rebus solving?" and something about using language puzzles for better understanding LLM abilities}

\section{Background and Related Work}

\paragraph{Italian \textit{Enigmistica} and Rebuses} The Italian language is characterized by a rich and long-standing tradition of puzzle games, including rebuses, dating back to the 19th century~\citep{enimmistica}\footnote{Refer to~\citet{miola-rebus,bartezzaghi-semiotica,loradesiatavola} for a comprehensive overview of peculiarities and norms in modern Italian rebuses.} In Italian rebuses, a \textbf{first pass} (\textit{prima lettura}) representing an intermediate solution of the puzzle is produced by combining graphemes with underlying image elements in a left-to-right direction (\Cref{fig:example}). Then, the letters and words of the first pass undergo a re-segmentation (\textit{cesura}) according to a \textbf{solution key} (\textit{chiave di lettura}\footnote{Referred to as \textit{diagramma} in jargon.}), which specifies the length of words in the \textbf{solution} (\textit{frase risolutiva}). The \textbf{verbalized rebuses} we introduce in this work are variants of textual rebuses (\textit{rebus descritto} or \textit{verbis}), where the text-based puzzle is crafted by replacing first pass words with their crossword definitions in a templated format (\Cref{fig:example}).

\paragraph{Linguistic Puzzles as NLP Progress Metrics} Language games have recently been adopted as challenging tasks for LLM evaluation~\citep{cryptic-crosswords,manna-etal-2024-riddle,puzzle-survey}. While works in this area have historically focused on English crosswords~\citep{probabilistic-crossword-solving,webcrow,english-crossword,cryptic-crosswords-v2}, recent tests focus on a more diverse set of games such as the New York Times' ``Connections''~\citep{connections} and ``Wordle''~\citep{wordle}. Automatic crossword solvers were also developed for  French~\citep{french-crossword},  German~\citep{german-crossword} and Italian~\citep{italian-crossword-solver,zugarini-etal-2024-clue-instruct}, while didactic crossword generators are available for Italian~\citep{italian-crossword-generation} and Turkish~\citep{turkish-crossword}. Relatedly, the Italian evaluation campaign EVALITA\footnote{\url{https://www.evalita.it}} recently hosted two shared tasks focusing on the word-guessing game ``La Ghigliottina'' (\textit{The Guillotine})~\citep{ghigliottinai-evalita,ghighliottina-original}. To our knowledge, our work is the first to attempt the computational modeling and evaluation of rebus-solving systems. Importantly, language games such as rebuses are not easily translatable into other languages due to their structural and cultural elements. This makes them a scarce but valuable resource for language-specific evaluations of language processing systems.

\paragraph{LLMs as Sequential Reasoners} State-of-the-art LLMs were shown to struggle to follow sequential instructions presented in a single query~\citep{sifo-benchmark}, but their performances improved significantly with ad-hoc training~\citep{sequential-instructions-ft}. This acts as an initial motivation for our rebus-solving fine-tuning experiments. In our evaluation, we also adopt few-shot prompting~\citep{gpt3} and chain-of-thought reasoning~\citep{cot}, which were both shown to strongly improve LLMs' abilities when solving complex multi-step tasks.

\section{Experimental Setup}

\paragraph{Data} We begin by extracting all rebuses' first passes and solutions available on Eureka5\footnote{\url{http://www.eureka5.it}, additional details in~\Cref{app:data}. Rebus illustrations are not available in Eureka5.}, an online repository of Italian puzzles. We refer to the resulting dataset containing 223k unique rebuses sourced from various publications as \textsc{EurekaRebus}. For crossword definitions, we use~\textsc{ItaCW}~\citep{italian-crossword-generation}, containing 125k unique definition-word pairs. We select only \textsc{EurekaRebus} examples in which all first pass words match an existing~\textsc{ItaCW} definition to enable verbalization, maintaining 83,157 examples for our modeling experiments.\footnote{Since verbalized rebus are produced from textual contents only, crossword definitions may refer to different word meanings (e.g. [Two soccer players] is used to represent the word ``wings'' in~\Cref{fig:example} despite not matching the word sense ``bird wings'' of the original image. This does not affect the validity of our task.} Since several~\textsc{ItaCW} words are associated with multiple definitions, we randomly sample definitions to promote diversity in the resulting verbalized rebuses. A test set of 2k examples\footnote{Composed by Test \textsc{id} and Test \textsc{ood}, described in~\Cref{sec:ood}} is kept aside for evaluation, and the remaining 81k examples are used for model training. 

\begin{table}
\small
\centering
\begin{tabular}{p{0.95\linewidth}}
\toprule
\textbf{Prompt} \\
\midrule
\textit{Risolvi gli indizi tra parentesi per ottenere una prima lettura, e usa la chiave di lettura per ottenere la soluzione del rebus.} \\[2em]
\textit{\underline{Rebus:} U [Lo è il passacavallo] LO [È fatta di vimini] F F [Decimi di chilo] S [Disusato soprabito] A [Un rampicante dei Tropici]} \\[3em]
\textit{\underline{Chiave di lettura:} 3 6 12 8} \\[4pt]
\end{tabular}
\begin{tabular}{p{0.18\linewidth}p{0.68\linewidth}}
\midrule
\textbf{Generation} & \textit{Procediamo alla risoluzione del rebus passo per passo:} \\[2em]

\multirow{10}{*}{\shortstack{Definition \\ resolution}} & \hspace{-1em}\ldelim\{{10}{*}[${}$] \textit{- U = U} \\
& \textit{- [Lo è il passacavallo] = nave} \\
& \textit{- L O = L O} \\
& \textit{- [È fatta di vimini] = cesta} \\
& \textit{- F F = F F} \\
& \textit{- [Decimi di chilo] = etti} \\
& \textit{- S = S} \\
& \textit{- [Disusato soprabito] = tait} \\
& \textit{- A = A} \\
& \textit{- [Un rampicante dei Tropici] = liana} \\[1em]
\multirow{2}{*}{First pass} & \hspace{-1em}\ldelim\{{2}{*}[${}$] \textit{\underline{Prima lettura:} U nave LO cesta F F etti S tait A liana} \\[2em]
& \textit{Ora componiamo la soluzione seguendo la chiave risolutiva:} \\[2em]
\multirow{4}{*}{\shortstack{Solution \\ segmen- \\ tation}} & \hspace{-1em}\ldelim\{{4}{*}[${}$] \textit{3 = Una} \\
& \textit{6 = veloce} \\
& \textit{12 = staffettista} \\
& \textit{8 = italiana} \\[1em]
\multirow{1}{*}{Solution} & \hspace{-1em}\ldelim\{{1}{*}[${}$] \textit{\underline{Soluzione:} Una veloce staffettista italiana} \\
\bottomrule
\end{tabular}
\caption{Template used in fine-tuning and prompting experiments with highlighted reasoning stages. Example rebus by \textit{Il Piacentino, Settimana Enigmistica n. 2942}
}
\vspace{-10pt}
\label{tab:example}
\end{table}

\paragraph{Models} We fine-tune Phi-3 Mini 3.8B 4K~\citep{phi3}, the most capable LLM below 4B parameters for a wide range of Italian language tasks\footnote{\url{https://hf.co/spaces/FinancialSupport/open_ita_llm_leaderboard}}. We use quantized low-rank adapters (QLoRA; \citealp{lora,qlora}) for efficient fine-tuning with Unsloth\footnote{\url{https://github.com/unslothai/unsloth}} and Transformers~\cite{wolf-etal-2020-transformers}, training the model for 5,000 steps with a batch size of 16 over 81k examples. For comparing our model performances, we select GPT-4o~\citep{gpt4o} and Claude-3.5 Sonnet~\citep{claude} as the current state-of-the-art for proprietary LLMs and the instruction-tuned variants of Qwen-2 72B~\citep{qwen2} and LLaMA-3 70B~\citep{llama3} %\footnote{\url{https://ai.meta.com/blog/meta-llama-3/}}
as the best-performing open-source LLMs according to the Invalsi Italian benchmark~\citep{invalsi}. These four systems are used as untrained baselines thanks to their instruction-following abilities and prompted for rebus solving in a few-shot setting. 

\paragraph{Format} \Cref{tab:example} presents an example in the templated format used for fine-tuning Phi-3.\footnote{An English example is available in~\Cref{tab:example-eng}} The model is prompted to reason step-by-step by 1) solving crossword definitions sequentially (\textbf{definition resolution}); 2) producing a \textbf{first pass} copying letters and definitions' words; 3) re-segmenting it into solution words based on the solution key (\textbf{solution segmentation}); and finally 4) producing the \textbf{solution} by copying re-segmented words. We automatically convert rebuses in this format by deriving the solution key from solution word lengths and dynamically infilling the available information into the template. We use a similar format for prompting experiments, with five in-context step-by-step demonstrations and an explicit instruction asking the model to stick to the previous examples' format to streamline solution parsing.

\paragraph{Metrics} For our granular evaluation of rebus-solving performance, we adopt the following set of metrics focusing on the first passes (FP) and solutions (S) generated by LLMs:

\begin{itemize}
    \item \textbf{Definition (Def.)}: Proportion of correctly guessed words during definition resolution.
    \item \textbf{First Pass Words/Letter Accuracy}: Proportion of correct words and letters in the generated first pass. Lower scores may indicate issues with assembling a first pass from previous information.
    \item \textbf{First Pass Exact Match (EM)}: Proportion of generated first passes matching the gold reference.
    \item \textbf{Solution Key Match}: Proportion of generated solution words matching the lengths specified by the solution key. Lower scores may indicate difficulty in respecting the given length constraints.
    \item \textbf{Solution First Pass Match}: Proportion of first pass characters employed to construct solution words. Lower scores indicate issues with using generated first pass characters in the solution.\footnote{In practice, we define this as $1 - \text{CER}(\text{FP}, \text{S})$, where CER is the character error rate~\citep{cer} between the two sequences (lowercased, whitespace removed) computed with \href{https://github.com/jitsi/jiwer}{Jiwer}}
    \item \textbf{Solution Words Accuracy}: Proportion of correct words in the generated solution.
    \item \textbf{Solution Exact Match (EM)}: Proportion of generated solutions matching the gold reference.
\end{itemize}

\section{Results}

\Cref{tab:results} presents our evaluation results. We observe that \textit{all prompted models perform poorly on the task}, with the overall best prompted system (Claude 3.5 Sonnet) obtaining the correct solution only for 24\% of the 2k tested examples. Notably, open-source systems perform significantly worse than proprietary ones, producing correct first passes only for 4\% of the examples, and next to no correct solutions. Our fine-tuned system largely outperforms all state-of-the-art prompted models, predicting the correct solution in 51\% of cases. From first pass metrics, it is evident these results can be largely explained by the poor word-guessing capabilities of the models, which are greatly improved with fine-tuning. For prompted models, the slight decrease in scores between Def. and FP Words also highlights issues with copying predicted words in the expected format. Finally, we observe that fine-tuning strongly improves the constraint-following abilities of our system, with prompted systems being less strict with applying length and letter-choice constraints for their solutions (Key/FP Match).

\begin{table*}[!h]
    \small
    %\centering
    \begin{tabular}{ll|cccc|cccc}
    \toprule
    \multirow{2}{*}{\textbf{Model}} & \multirow{2}{*}{\textbf{Setup}} & \multirow{2}{*}{\textbf{Def.}} & \multicolumn{3}{c}{\textbf{First Pass (FP)}} & \multicolumn{4}{c}{\textbf{Solution (S)}} \\
    \cmidrule(lr){4-6}
    \cmidrule(lr){7-10}
    & & & Words & Letters & EM & Key Match & FP Match & Words & EM \\
    \midrule
    LLaMA-3 70B         & 5-shot prompt & 0.22 & 0.20 & 0.60 & 0.04 & 0.16 & 0.51 & 0.03 & 0.00\\
    Qwen-2 72B          & 5-shot prompt & 0.28 & 0.25 & 0.76 & 0.04 & 0.20 & 0.52 & 0.04 & 0.00 \\
    GPT-4o              & 5-shot prompt & 0.55 & 0.51 & 0.83 & 0.15 & 0.53 & 0.74 & 0.27 & 0.11 \\
    Claude-3.5 Sonnet   & 5-shot prompt & \underline{0.66} & \underline{0.62} & \underline{0.90} & \underline{0.28} & \underline{0.83} & \underline{0.82} & \underline{0.43} & \underline{0.24} \\
    \midrule
    Phi-3 3.8B (ours) & fine-tuned & \textbf{0.84} & \textbf{0.84} & \textbf{1.00} & \textbf{0.56} & \textbf{0.86} & \textbf{0.94} & \textbf{0.68} & \textbf{0.51} \\
    \bottomrule
    \end{tabular}
    \caption{Fine-grained verbalized rebus solving performances of various LLMs. \textbf{Bold} denotes best overall performances, and \underline{underline} marks best training-free results.}
    \label{tab:results}
    \vspace{-10pt}
\end{table*}

\section{What Motivates Model Performances?}

In light of the strong performances achieved by our relatively small fine-tuned system, this section conducts an in-depth investigation to identify factors motivating such performance improvements.

\paragraph{Word Complexity and Frequency Affects LLM Fine-tuning Performance}
\label{sec:corr} For every word in the first passes and solutions of test set examples, we measure LLMs' overall accuracy in predicting it for the full test set. We then correlate this score to various quantities that could motivate LLMs' performances. More specifically, we use 1) the word frequency in the training set; 2) the word frequency in \textsc{Paisà}~\citep{paisa}, a large web Italian corpus; and 3) the length of the word (number of characters). We find a significant positive correlation ($\rho = 0.44$) between first pass word prediction accuracy and training frequency for the fine-tuned Phi-3 model, suggesting that model performance is strongly related to training coverage. The length of characters is also found to negatively affect our model's performance, albeit to a smaller extent ($\rho = -0.11$). The performance of prompted models is unrelated to both properties for first pass words, indicating that these results are the product of fine-tuning.\footnote{\textsc{Paisà} frequency is never found to correlate significantly. Full correlation results are available in~\Cref{tab:correlations}.}

\paragraph{LLM Fine-Tuning Fails to Generalize to Unseen Words}
\label{sec:ood}

\begin{table}
    \small
    \begin{tabular}{l|ccc|ccc}
    \toprule
    \multirow{2}{*}{\textbf{Metric}} & \multicolumn{3}{c}{\textbf{GPT-4o}} & \multicolumn{3}{c}{\textbf{Phi-3 (ours)}} \\
    \cmidrule(lr){2-4}
    \cmidrule(lr){5-7}
    & \shortstack{\textbf{Test} \\ \textbf{\textsc{id}}} & \shortstack{\textbf{Test} \\ \textbf{\textsc{ood}}} & \shortstack{\textbf{Test} \\ $\mathbf{\Delta}$} & \shortstack{\textbf{Test} \\ \textbf{\textsc{id}}} & \shortstack{\textbf{Test} \\ \textbf{\textsc{ood}}} & \shortstack{\textbf{Test} \\ $\mathbf{\Delta}$} \\
     \midrule
     FP W.$~_\textsc{ID}$    & 0.52 & 0.51 & -0.01 & 0.96 & 0.96 & 0.00 \\
     FP W.$~_\textsc{OOD}$   & - & 0.44 & - & -    & 0.20 & -    \\
     FP EM                  & 0.16 & 0.14 & -0.02 & 0.89 & 0.18 & \textcolor{red}{-0.71} \\
     \midrule
     S W.$~_\textsc{ID}$     & 0.29 & 0.26 & -0.03 & 0.92 & 0.49 & \textcolor{red}{-0.43} \\
     S W.$~_\textsc{OOD}$    & 0.18 & 0.16 & -0.02 & 0.63 & 0.20 & \textcolor{red}{-0.40} \\
     S EM                   & 0.12 & 0.09 & -0.03 & 0.82 & 0.16 & \textcolor{red}{-0.66} \\
    \bottomrule
    \end{tabular}
    \caption{Model performances for test subsets containing only in-domain (Test ID), or some out-of-domain (Test OOD) first pass words. W.$~_\textsc{ID}$ and W.$~_\textsc{OOD}$ are accuracies for ID and OOD words for first pass (FP) and solution (S) sequences. Test $\Delta$ = Test ID - Test OOD performance.}
    \label{tab:ood}
    \vspace{-10pt}
\end{table}

\begin{table}
\small
\centering
\begin{tabular}{p{0.95\linewidth}}
\toprule
\textit{\underline{Rebus:} SAP [La porta della breccia]$^{~\text{D}1}$ TE [La pinza del granchio]$^{~\text{D}2}$ SBA [Si legge su alcuni orologi]$^{~\text{D}3}$ G [Le sue coccole sono aromatiche]$^{~\text{D}4}$ V [Un gioco con dadi e pedine]$^{~\text{D}5}$ D [Sono verdi in gioventù]$^{~\text{D}6}$} \\[2em]
\textit{\underline{Chiave di lettura:} \textcolor{blue}{\textbf{8 3 2 12 7 5}}} \\[4pt]
\midrule
\end{tabular}
\begin{tabular}{lccc}
\textbf{Step} & \textbf{GPT-4o} & \textbf{Claude 3.5S} & \textbf{Phi-3} \\
D1 & \textcolor{Red}{p}         & \textcolor{Red}{one}         & \textcolor{Green}{pia}     \\
D2 & \textcolor{Green}{chela}   & \textcolor{Green}{chela}   & \textcolor{Green}{chela}   \\
D3 & \textcolor{Red}{ora}       & \textcolor{Green}{data}       & \textcolor{Green}{data}    \\
D4 & \textcolor{Green}{ginepro} & \textcolor{Red}{lio} & \textcolor{Green}{ginepro} \\
D5 & \textcolor{Red}{ludo}      & \textcolor{Green}{oca}      & \textcolor{Green}{oca}     \\
D6 & \textcolor{Red}{acerbi}    & \textcolor{Green}{anni}    & \textcolor{Green}{anni}    \\
\midrule
S\textcolor{blue}{\textbf{8}}  & \textcolor{Red}{Spettacolo}     & \textcolor{Red}{Saponate} & \textcolor{Green}{Sappiate}     \\
S\textcolor{blue}{\textbf{3}}  & \textcolor{Green}{che}          & \textcolor{Green}{che}      & \textcolor{Green}{che}   \\
S\textcolor{blue}{\textbf{2}}  & \textcolor{Red}{fa}             & \textcolor{Green}{la}         & \textcolor{Green}{la}    \\
S\textcolor{blue}{\textbf{12}} & \textcolor{Red}{sognare}        & \textcolor{Green}{sbadataggine} & \textcolor{Green}{sbadataggine} \\
S\textcolor{blue}{\textbf{7}}  & \textcolor{Red}{ogni}           & \textcolor{Red}{vocando}    & \textcolor{Green}{provoca}     \\
S\textcolor{blue}{\textbf{5}}  & \textcolor{Red}{sera}           & \textcolor{Green}{danni}       & \textcolor{Green}{danni}    \\
\midrule
\multicolumn{4}{c}{\textit{\underline{Soluzione:} SAP\textcolor{Green}{pia}TE \textcolor{Green}{che la} SBA\textcolor{Green}{data}G\textcolor{Green}{gine pro}V\textcolor{Green}{oca} D\textcolor{Green}{anni}}} \\
\end{tabular}
\begin{tabular}{p{0.95\linewidth}}
\toprule
\toprule
\textit{\underline{Rebus:} STU [Si salva otturandolo]$^{~\text{D}1}$ S [Ha foglie seghettate]$^{~\text{D}2}$ AL [Lo è l'operaio che lavora in cantiere]$^{~\text{D}3}$ G [Un uomo... non all' altezza]$^{~\text{D}4}$} \\[2em]
\textit{\underline{Chiave di lettura:} \textcolor{blue}{\textbf{11 7 2 7}}} \\[4pt]
\midrule
\end{tabular}
\begin{tabular}{lccc}
\textbf{Step} & \textbf{GPT-4o} & \textbf{Claude 3.5S} & \textbf{Phi-3} \\
D1 & \textcolor{Red}{tappo}         & \textcolor{Red}{falla}         & \textcolor{Green}{dente}     \\
D2 & \textcolor{Red}{acero}   & \textcolor{Red}{ortica}   & \textcolor{Red}{aro}   \\
D3 & \textcolor{Green}{edile}       & \textcolor{Green}{edile}       & \textcolor{Green}{edile}    \\
D4 & \textcolor{Green}{nano} & \textcolor{Green}{nano} & \textcolor{Green}{nano} \\
\midrule
S\textcolor{blue}{\textbf{11}}  & \textcolor{Red}{Stupaccerone}     & \textcolor{Red}{Stufallassor} & \textcolor{Red}{Studentesaro}     \\
S\textcolor{blue}{\textbf{7}}  & \textcolor{Red}{salendo}          & \textcolor{Red}{ticale}      & \textcolor{Red}{aledile}   \\
S\textcolor{blue}{\textbf{2}}  & \textcolor{Red}{al}             & \textcolor{Green}{di}         & \textcolor{Red}{gi}    \\
S\textcolor{blue}{\textbf{7}} & \textcolor{Red}{genano}        & \textcolor{Green}{Legnano} & \textcolor{Red}{nanano} \\
\midrule
\multicolumn{4}{c}{\textit{\underline{Soluzione:} STU\textcolor{Green}{dente}S\textcolor{Green}{sa lice}AL\textcolor{Green}{e di Le}G\textcolor{Green}{nano}}} \\
\end{tabular}
\caption{Examples of LLM generations for rebuses by \textit{Slam, Nuova Enigmistica Tascabile n. 2802} (top) and \textit{Grizzly, Domenica Quiz n. 2} (bottom). \textcolor{Green}{Correct guesses} and \textcolor{Red}{errors} and denoted for predicted first pass definitions (D$_{1,\dots,N}$) and solution words (S$_i$, with $i$ being the $i$-th solution key value).}
\vspace{-10pt}
\label{tab:failures}
\end{table}

To further confirm the importance of fine-tuning word coverage in defining model performances, we evaluate our fine-tuned model in out-of-distribution settings. For this evaluation, the 2k examples of the test set from previous sections are divided into two subsets: one in which all first pass words were seen during fine-tuning by Phi-3 (\textbf{Test \textsc{id}}, 1061 examples) and one in which, for every example, at least one first pass word was unseen in training (\textbf{Test \textsc{ood}}, 939 examples). Intuitively, if Phi-3 performance is mainly motivated by memorizing fine-tuning data, introducing OOD words should produce a significant drop in model performances. Results shown in~\Cref{tab:ood} confirm that this is indeed the case. We find Phi-3 performances to be near-perfect on seen first pass words (FP W.$~_\textsc{ID}$ = 0.96) in both test sets, with a major drop for OOD words (FP W.$~_\textsc{OOD}$ = 0.20). This produces second-order effects on subsequent steps, causing the FP EM results to drop by 71\% (FP EM Test $\Delta$), while significantly impacting downstream solution accuracies. On the contrary, GPT-4o few-shot prompting performances remain nearly identical on both splits, confirming that these results are not the product of a skewed data selection process. Overall, these results strongly suggest that memorization is the main factor behind the strong rebus-solving performance of our fine-tuned LLM.

\paragraph{Manual Inspection}

We conclude by manually evaluating some generations produced by the best-performing LLMs. \Cref{tab:failures} presents two examples with definitions (D) and solution (S) words predicted by three LLMs, with more examples provided in~\Cref{app:more-examples}. We use \textsc{naw} as short-hand for ``Not A Word'' to mark nonsensical terms.

In the first example, Phi-3 correctly predicts all first pass and solution words. On the contrary, other models make several mistakes in the first pass, leading to incorrect solutions. Both prompted models tend to ignore first pass words when these cannot be assembled to form sensical, length-fitting solution words. For example, for D1 GPT-4o predicts \textcolor{Red}{p} (\textsc{naw}), which would lead to the solution word ``SAP\textcolor{Red}{p}TE'' (\textsc{naw}), but the S8 = ``\textcolor{Red}{Spettacolo}'' (\textit{show}) is predicted instead by the model). In particular, GPT-4o appears to prioritize grammatically correct solutions at the cost of ignoring first pass words and solution key length constraints, while Claude 3.5S shows an improved ability to follow these constraints, as confirmed by Key/FP Match results of~\Cref{tab:results}.

In the second example, the first pass word D2 = \textcolor{Green}{salice} (\textit{willow}) is OOD for Phi-3. Consequently, the model produces the incorrect prediction \textcolor{Red}{aro} (\textsc{naw}), and the error is propagated to all solution words, as previously observed in the Test OOD column of~\Cref{tab:ood}. Prompted models also underperform in this example, with errors on D1 and D2 propagating to most solution words. However, we note that D1 and D2 incorrect predictions for Claude 3.5S satisfy the provided definitions, suggesting that access to more explicit information about the given constraints could further boost LLMs' performance on this task.

\section{Discussion and Conclusion}

This work introduced a verbalized rebus-solving task and dataset for evaluating LLMs' sequential instruction following skills for the Italian language. We crafted a large collection of 83k verbalized rebuses by combining rebus transcriptions with crossword definitions and used it to evaluate the rebus-solving skills of state-of-the-art LLMs. Our experiments revealed the challenging nature of this task, with even the most capable prompted models achieving only 24\% accuracy on solutions.

While fine-tuning a smaller LLM dramatically improved performance to 51\% solution accuracy, our analysis uncovered that these gains were largely driven by memorization and do not generalize to out-of-distribution examples. These results suggest important limitations in the generalization capabilities of current systems for sequential instruction following tasks. Our manual analysis further shows that LLMs seldom account for length constraints when solving definitions, despite the fundamental role of these cues in restricting the pool of possible words. These results suggest that search-based approaches accounting for constraints more explicitly might improve puzzle structure adherence, as previously shown by~\citet{crossword-mcts}. Other augmentation techniques employing LLM reformulation skills can also be explored to mitigate overfitting.

Future work in this area should focus on expanding similar evaluations to a wider set of languages, input modalities, and puzzle categories, creating a comprehensive benchmark to test LLMs' puzzle-solving skills. Importantly, the task of solving visual rebuses and their more convoluted variants\footnote{For example, rebuses requiring first pass anagrams (\textit{anarebus}) or dynamic relations derived from multi-scene analysis  (\textit{stereorebus})} remains far beyond the current capabilities of vision-language models. Hence, solving these puzzles automatically can be considered an important milestone in developing multimodal AI systems for constrained multi-step reasoning tasks. Our results confirm that the challenging nature of rebuses, even in their verbalized form, makes this task valuable for assessing future progress in LLMs' linguistic proficiency and sequential reasoning abilities. Finally, our rebus-solving LLM can facilitate future interpretability work investigating the mechanisms behind factual recall and multi-step reasoning in transformer models~\citep{ferrando-etal-2024-primer}.

\paragraph{Limitations} 
Our analysis was limited to a relatively small set of models, and a single prompt template obtained after minimal tuning. Further experiments are needed to verify that memorization patterns after fine-tuning remain relevant for other model sizes, prompt formats, and training regimes, particularly for full-weight training approaches.

\begin{acknowledgments}
   Gabriele Sarti and Arianna Bisazza acknowledge the support of the Dutch Research Council (NWO) for the project InDeep (NWA.1292.19.399). 
   Arianna Bisazza is further supported by the NWO Talent Programme (VI.Vidi.221C.009).
   We are grateful to the \href{http://www.enignet.it/home}{Associazione Culturale ``Biblioteca Enigmistica Italiana - G. Panini''} for making its rebus collection freely accessible on the Eureka5 platform, and to Valeriya Zelenkova for her valuable comments on the first version of this work.
\end{acknowledgments}

\bibliography{bibliography}

\appendix

\section{Additional Data Information}
\label{app:data}

\paragraph{Dataset statistics} \Cref{tab:stats} presents statistics for the \textsc{EurekaRebus} dataset and the filtered subset we use for composing verbalized rebuses. The \textsc{ItaCW} dataset contains a total of 125,202 definitions for 40,963 unique words, with the most frequent words having hundreds of different definitions, e.g. 173 for \textit{re} (king), 155 for \textit{te} (you). Definitions used for verbalization are randomly sampled from the pool of available definitions for every word.

\begin{figure*}
    \centering
    \includegraphics[width=\textwidth]{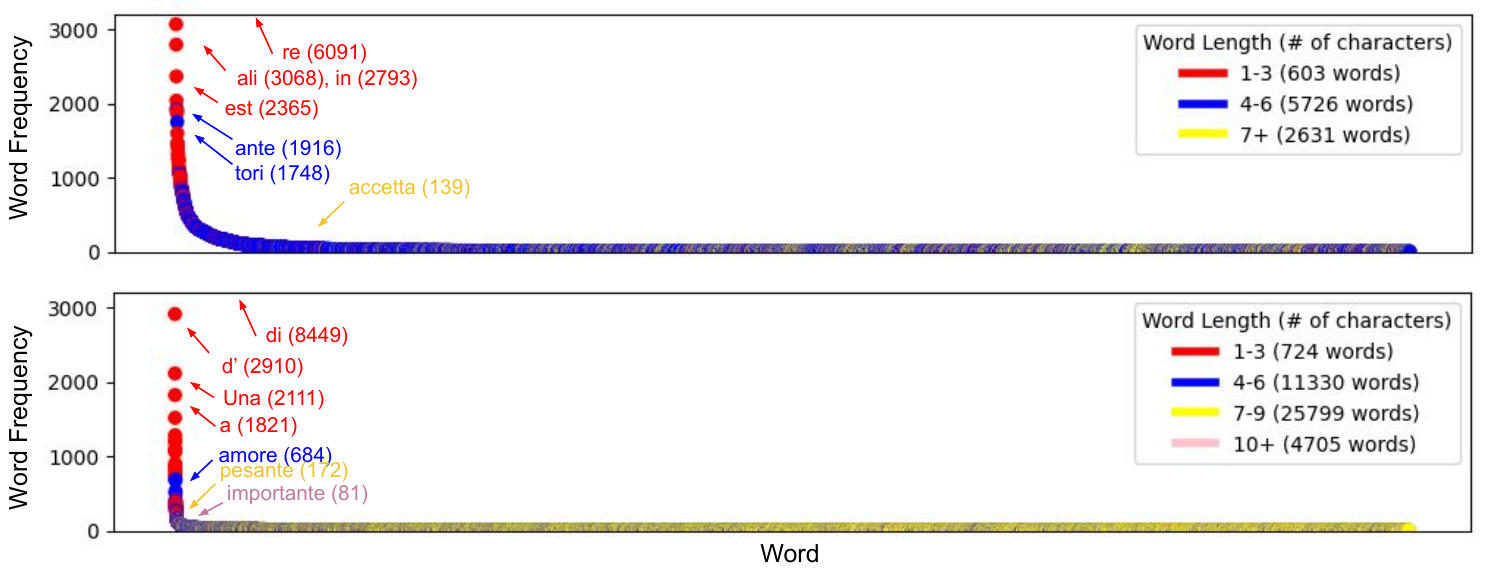}
    \caption{Word frequencies for words in first passes (top) and solutions (bottom) for the selected subset of \textsc{EurekaRebus} used for training and evaluation. Words are colored according to their length, and the most frequent examples per frequency bin are highlighted.}
    \label{fig:distr}
\end{figure*}

\begin{table}
    \small
    \begin{tabular}{l|cc}
    \toprule
    \textbf{Statistic} & \textbf{\textsc{EurekaRebus}} & \textbf{\textsc{ItaCW}-filtered} \\
    \midrule
    \# examples        & 222089      & 83157 \\
    \# authors         & 8138        & 5046 \\
    Year range         & 1800 - 2024 & 1869 - 2024 \\
    \midrule
    \multicolumn{3}{c}{\textbf{First pass}} \\
    \midrule
    \# unique words    & 38977       & 8960       \\
    Avg./SD words/ex.  & 3.50/1/48   & 3.08/1.00  \\
    Avg./SD word len.  & 6.51/1.96   & 5.70/1.60  \\
    Avg./SD FP len.    & 26.45/11.19 & 25.74/8.73 \\
    \midrule
    \multicolumn{3}{c}{\textbf{Solution}} \\
    \midrule
    \# unique words    & 75718      & 42558 \\
    Avg./SD words/ex.  & 3.02/1.60  & 2.80/1.21  \\
    Avg./SD word len.  & 8.07/2.30  & 7.79/2.23  \\
    Avg./SD Sol. len.    & 19.47/8.44 & 18.81/6.06 \\
    \bottomrule
    \end{tabular}
    \caption{Statistics for the full \textsc{EurekaRebus} dataset and the crosswords-filtered subset used in this work. Avg./SD = Average/standard deviation.}
    \label{tab:stats}
\end{table}

\paragraph{First pass/Solution word distribution} \Cref{fig:distr} shows the distribution of first pass and solution words for the filtered \textsc{EurekaRebus} subset used in our work.

\section{Additional Experimental Results}
\label{app:corr}

\Cref{tab:correlations} presents the correlations between model accuracy and the properties presented in~\Cref{sec:corr}. \Cref{tab:ood-full} presents the full ID/OOD performances for all tested models, showing consistent results with~\Cref{tab:ood} for all prompted models.~\Cref{tab:train-results} presents Phi-3 Mini performances across rebus-solving fine-tuning steps.

\begin{table}
    \small
    \begin{tabular}{l|ccc}
    \toprule
    \textbf{Model} & \textbf{\# Char.} & \textbf{Paisà Freq.} & \textbf{Train Freq.} \\
    \midrule
    GPT-4o      & -0.01 & 0.01 & 0.02 \\
    Claude-3.5 & -0.02 & -0.02 & 0.00 \\
    Phi-3 (ours) & \textbf{-0.11} & -0.05 & \textbf{0.44} \\
    \bottomrule
    \toprule
    GPT-4o      & \textbf{-0.18} & \textbf{0.14} & \textbf{0.19} \\
    Claude-3.5 & \textbf{-0.15} & \textbf{0.08} & \textbf{0.13} \\
    Phi-3 (ours) & -0.02 & \textbf{0.08} & \textbf{0.22} \\
    \bottomrule
    \end{tabular}
    \caption{Spearman's correlation with average word accuracies for metrics computed on first pass (top) and solution (bottom) words. \textbf{Bold scores} are significant with Bonferroni-corrected $p < 1e-5$~\citep{bonferroni}}
    \label{tab:correlations}
    \vspace{-10pt}
\end{table}

\begin{table*}
    \footnotesize
    \begin{tabular}{l|ccc|ccc|ccc|ccc|ccc}
    \toprule
    \multirow{2}{*}{\textbf{Metric}} & \multicolumn{3}{c}{\textbf{LLaMA-3}} & \multicolumn{3}{c}{\textbf{Qwen-2}} & \multicolumn{3}{c}{\textbf{GPT-4o}} & \multicolumn{3}{c}{\textbf{Claude-3.5S}} & \multicolumn{3}{c}{\textbf{Phi-3 (ours)}} \\
    \cmidrule(lr){2-4}
    \cmidrule(lr){5-7}
    \cmidrule(lr){8-10}
    \cmidrule(lr){11-13}
    \cmidrule(lr){14-16}
    & \shortstack{\textbf{Test} \\ \textbf{\textsc{id}}} & \shortstack{\textbf{Test} \\ \textbf{\textsc{ood}}} & \shortstack{\textbf{Test} \\ $\mathbf{\Delta}$} & \shortstack{\textbf{Test} \\ \textbf{\textsc{id}}} & \shortstack{\textbf{Test} \\ \textbf{\textsc{ood}}} & \shortstack{\textbf{Test} \\ $\mathbf{\Delta}$} & \shortstack{\textbf{Test} \\ \textbf{\textsc{id}}} & \shortstack{\textbf{Test} \\ \textbf{\textsc{ood}}} & \shortstack{\textbf{Test} \\ $\mathbf{\Delta}$} & \shortstack{\textbf{Test} \\ \textbf{\textsc{id}}} & \shortstack{\textbf{Test} \\ \textbf{\textsc{ood}}} & \shortstack{\textbf{Test} \\ $\mathbf{\Delta}$} & \shortstack{\textbf{Test} \\ \textbf{\textsc{id}}} & \shortstack{\textbf{Test} \\ \textbf{\textsc{ood}}} & \shortstack{\textbf{Test} \\ $\mathbf{\Delta}$}\\
     \midrule
     FP W.$~_\textsc{ID}$     & 0.20 & 0.19 & -0.01 & 0.26 & 0.25 & -0.01 & 0.52 & 0.51 & -0.01 & 0.65 & 0.63 & -0.02 & 0.96 & 0.96 & 0.00                   \\
     FP W.$~_\textsc{OOD}$    & -    & 0.18 & -     & -    & 0.24 & -     & -    & 0.44 & -     & -    & 0.54 & -     & -    & 0.20 & -                      \\
     FP EM                  & 0.03 & 0.04 & 0.01  & 0.03 & 0.05 & 0.02  & 0.16 & 0.14 & -0.02 & 0.30 & 0.25 & -0.05 & 0.89 & 0.18 & \textcolor{red}{-0.71} \\
     \midrule
     S W.$~_\textsc{ID}$      & 0.03 & 0.04 & 0.01  & 0.04 & 0.05 & 0.01  & 0.29 & 0.26 & -0.03 & 0.48 & 0.40 & -0.08 & 0.92 & 0.49 & \textcolor{red}{-0.43} \\
     S W.$~_\textsc{OOD}$     & 0.01 & 0.00 & -0.01 & 0.02 & 0.00 & -0.02 & 0.18 & 0.16 & -0.02 & 0.41 & 0.30 & -0.11& 0.63 & 0.20 & \textcolor{red}{-0.40} \\
     S EM                   & 0.00 & 0.00 & 0.00  & 0.00 & 0.00 & 0.00  & 0.12 & 0.09 & -0.03 & 0.27 & 0.22 & -0.05 & 0.82 & 0.16 & \textcolor{red}{-0.66} \\
    \bottomrule
    \end{tabular}
    \caption{Full model performances for test subsets containing only in-domain (Test ID), or some out-of-domain (Test OOD) first pass words. W.$~_\textsc{ID}$ and W.$~_\textsc{OOD}$ are accuracies for ID and OOD words for first pass (FP) and solution (S) sequences. Test $\Delta$ = Test ID - Test OOD performance.}
    \label{tab:ood-full}
    \vspace{-10pt}
\end{table*}

\begin{table*}
    \small
    %\centering
    \begin{tabular}{l|cccc|cccc}
    \toprule
    \multirow{2}{*}{\textbf{\# Train Steps}} & \multirow{2}{*}{\textbf{Def.}} & \multicolumn{3}{c}{\textbf{First Pass (FP)}} & \multicolumn{4}{c}{\textbf{Solution (S)}} \\
    \cmidrule(lr){3-5}
    \cmidrule(lr){6-9}
    & & Words & Letters & EM & Key Match & FP Match & Words & EM \\
    \midrule
     500 & 0.64 & 0.63 & 0.97 & 0.25 & 0.66 & 0.86 & 0.36 & 0.16 \\
    1000 & 0.74 & 0.74 & 1.00 & 0.38 & 0.72 & 0.89 & 0.48 & 0.28 \\
    1500 & 0.78 & 0.77 & 0.99 & 0.42 & 0.78 & 0.91 & 0.55 & 0.34 \\
    2000 & 0.80 & 0.79 & 1.00 & 0.47 & 0.81 & 0.93 & 0.59 & 0.40 \\
    2500 & 0.81 & 0.81 & 1.00 & 0.49 & 0.81 & 0.92 & 0.62 & 0.42 \\
    3000 & 0.82 & 0.82 & 1.00 & 0.51 & 0.82 & 0.92 & 0.63 & 0.44 \\
    3500 & 0.82 & 0.82 & 1.00 & 0.53 & 0.83 & 0.93 & 0.65 & 0.47 \\
    4000 & 0.83 & 0.83 & 1.00 & 0.53 & 0.85 & 0.93 & 0.66 & 0.47 \\
    4500 & 0.83 & 0.83 & 1.00 & 0.55 & 0.85 & 0.93 & 0.67 & 0.49 \\
    5072 (final) & \textbf{0.84} & \textbf{0.84} & \textbf{1.00} & \textbf{0.56} & \textbf{0.86} & \textbf{0.94} & \textbf{0.68} & \textbf{0.51} \\
    \bottomrule
    \end{tabular}
    \caption{Fine-grained verbalized rebus solving performances of the Phi-3 Mini model on the test set including both ID and OOD examples at various fine-tuned stages. \textbf{Bold} denotes best overall performances.}
    \label{tab:train-results}
    \vspace{-10pt}
\end{table*}

\section{Additional Model Generations}
\label{app:more-examples}

\Cref{tab:example-eng} presents an English translation of~\Cref{fig:example} example using the prompt format adopted in this study. \Cref{tab:more-ex-1,tab:more-ex-2} provide additional example of LLM generations for tested rebuses, with the example from~\Cref{tab:more-ex-2} (bottom) being OOD due to the \textit{manovella} (crank) word in D2, and the others being ID for the fine-tuned Phi-3 Mini.

\begin{table}
\small
\centering
\begin{tabular}{p{0.95\linewidth}}
\toprule
\textbf{Prompt} \\
\midrule
\textit{Solve the clues provided between parentheses to obtain a first pass, and use the solution key to obtain the rebus' solution.} \\[2em]
\textit{\underline{Rebus:} M [Two attacking footballers] N [Used for eating icecream] [Barks and bites] NIA} \\[2em]
\textit{\underline{Solution key:} 11 5} \\[4pt]
\end{tabular}
\begin{tabular}{p{0.18\linewidth}p{0.68\linewidth}}
\midrule
\textbf{Generation} & \textit{Let's proceed to solve the rebus step by step:} \\[1em]

\multirow{6}{*}{\shortstack{Definition \\ resolution}} & \hspace{-1em}\ldelim\{{6}{*}[${}$] \textit{- M = M} \\
& \textit{- [Two attacking footballers] =} ali \textit{(wings)} \\
& \textit{- N = N} \\
& \textit{- [Used for eating icecream] =} coni \textit{(cones)} \\
& \textit{- [Barks and bites] =} cane \textit{(dog)} \\
& \textit{- N I A = N I A} \\[1em]
\multirow{1}{*}{First pass} & \hspace{-1em}\ldelim\{{1}{*}[${}$] \textit{\underline{First pass:} M ali N coni cane NIA} \\[1em]
& \textit{Let's now compose the solution by following the solution key:} \\[2em]
\multirow{2}{*}{\shortstack{Solution \\ segmen- \\ tation}} & \hspace{-1em}\ldelim\{{2}{*}[${}$] \textit{11 =} Malinconica \textit{(melancholic)} \\
& \textit{5 =} nenia \textit{(lullaby)} \\[1em]
\multirow{2}{*}{Solution} & \hspace{-1em}\ldelim\{{2}{*}[${}$] \textit{\underline{Solution:}} Malinconica nenia \textit{(melancholic lullaby)} \\
\bottomrule
\end{tabular}
\caption{English translation of the template used in fine-tuning and prompting experiments with highlighted reasoning stages, with translated first pass and solution words in parentheses. Example rebus by \textit{A. Rizzi, Settimana Enigmistica n. 4656}
}
\vspace{-10pt}
\label{tab:example-eng}
\end{table}

\begin{table}
\small
\centering
\begin{tabular}{p{0.95\linewidth}}
\toprule
\textit{\underline{Rebus:} MOR [Musa della commedia]$^{~\text{D}1}$ L [Si accendono per vedere]$^{~\text{D}2}$ NO [Uomini di intelligenza superiore]$^{~\text{D}3}$} \\[2em]
\textit{\underline{Chiave di lettura:} \textcolor{blue}{\textbf{7 12}}} \\[4pt]
\midrule
\end{tabular}
\begin{tabular}{lccc}
\textbf{Step} & \textbf{GPT-4o} & \textbf{Claude 3.5S} & \textbf{Phi-3} \\
D1 & \textcolor{Green}{Talia}         & \textcolor{Green}{Talia}         & \textcolor{Green}{Talia}     \\
D2 & \textcolor{Green}{luci}   & \textcolor{Green}{luci}   & \textcolor{Green}{luci}   \\
D3 & \textcolor{Green}{geni}       & \textcolor{Green}{geni}       & \textcolor{Red}{genii}    \\
\midrule
S\textcolor{blue}{\textbf{7}}  & \textcolor{Green}{Mortali}     & \textcolor{Green}{Mortali} & \textcolor{Green}{Mortali}     \\
S\textcolor{blue}{\textbf{12}}  & \textcolor{Green}{allucinogeni}          & \textcolor{Green}{allucinogeni}      & \textcolor{Green}{allucinogeni}   \\
\midrule
\multicolumn{4}{c}{\textit{\underline{Soluzione:} MOR\textcolor{Green}{Tali} \textcolor{Green}{a}L\textcolor{Green}{luci}NO\textcolor{Green}{geni}}} \\
\end{tabular}
\caption{Examples of LLM generations for a rebus by \textit{De Vico C., Domenica Quiz n. 5}. \textcolor{Green}{Correct guesses} and \textcolor{Red}{errors} and denoted for predicted first pass definitions (D$_{1,\dots,N}$) and solution words (S$_i$, with $i$ being the $i$-th solution key value).}
\vspace{-10pt}
\label{tab:more-ex-1}
\end{table}

\begin{table}
\small
\centering
\begin{tabular}{p{0.95\linewidth}}
\toprule
\textit{\underline{Rebus:} PRI [Ricoperto di sudore]$^{~\text{D}1}$ MIN [Gli altari del tempio]$^{~\text{D}2}$ DO [Un ordigno bellico]$^{~\text{D}3}$ G [ossono essere ``di serie'' in certi tornei]$^{~\text{D}4}$ SSO} \\[2em]
\textit{\underline{Chiave di lettura:} \textcolor{blue}{\textbf{5 2 8 6 2 6}}} \\[4pt]
\midrule
\end{tabular}
\begin{tabular}{lccc}
\textbf{Step} & \textbf{GPT-4o} & \textbf{Claude 3.5S} & \textbf{Phi-3} \\
D1 & \textcolor{Red}{sudato}         & \textcolor{Green}{madido}         & \textcolor{Green}{madido}     \\
D2 & \textcolor{Green}{are}   & \textcolor{Green}{are}   & \textcolor{Green}{are}   \\
D3 & \textcolor{Red}{bomba}       & \textcolor{Green}{mina}       & \textcolor{Green}{mina}    \\
D4 & \textcolor{Green}{teste} & \textcolor{Green}{teste} & \textcolor{Green}{teste} \\
\midrule
S\textcolor{blue}{\textbf{5}}  & \textcolor{Green}{Prima}     & \textcolor{Green}{Prima} & \textcolor{Green}{Prima}     \\
S\textcolor{blue}{\textbf{2}}  & \textcolor{Green}{di}          & \textcolor{Green}{di}      & \textcolor{Green}{di}   \\
S\textcolor{blue}{\textbf{8}}  & \textcolor{Red}{sudare}             & \textcolor{Red}{minaccia}         & \textcolor{Green}{dominare}    \\
S\textcolor{blue}{\textbf{6}} & \textcolor{Red}{molto}        & \textcolor{Red}{teste} & \textcolor{Red}{dominate} \\
S\textcolor{blue}{\textbf{2}}  & \textcolor{Red}{di}           & \textcolor{Red}{di}    & \textcolor{Red}{se}     \\
S\textcolor{blue}{\textbf{6}}  & \textcolor{Red}{testa}           & \textcolor{Red}{dosso}       & \textcolor{Green}{stesso}    \\
\midrule
\multicolumn{4}{c}{\textit{\underline{Soluzione:} PRI\textcolor{Green}{ma di do}MIN\textcolor{Green}{are} DO\textcolor{Green}{mina} \textcolor{Green}{te ste}SSO}} \\
\end{tabular}
\begin{tabular}{p{0.95\linewidth}}
\toprule
\toprule
\textit{\underline{Rebus:} AT [Si alzano nel camping]$^{~\text{D}1}$ [Emoziona pescatori e navigatori]$^{~\text{D}2}$ [Come una nota Foresta]$^{~\text{D}3}$ MEN [Quadro ad olio]$^{~\text{D}4}$ S [Atteggiamento da modella]$^{~\text{D}5}$} \\[2em]
\textit{\underline{Chiave di lettura:} \textcolor{blue}{\textbf{9 11 2 5}}} \\[4pt]
\midrule
\end{tabular}
\begin{tabular}{lccc}
\textbf{Step} & \textbf{GPT-4o} & \textbf{Claude 3.5S} & \textbf{Phi-3} \\
D1 & \textcolor{Green}{tende}         & \textcolor{Green}{tende}         & \textcolor{Green}{tende}     \\
D2 & \textcolor{Red}{marea}   & \textcolor{Red}{mare}   & \textcolor{Green}{rete}   \\
D3 & \textcolor{Green}{nera}       & \textcolor{Green}{nera}       & \textcolor{Green}{nera}    \\
D4 & \textcolor{Red}{dipinto} & \textcolor{Green}{tela} & \textcolor{Green}{tela} \\
D5 & \textcolor{Green}{posa} & \textcolor{Green}{posa} & \textcolor{Green}{posa} \\
\midrule
S\textcolor{blue}{\textbf{9}}  & \textcolor{Red}{Attenderemo}     & \textcolor{Green}{Attendere} & \textcolor{Green}{Attendere}     \\
S\textcolor{blue}{\textbf{11}}  & \textcolor{Red}{mareanera}          & \textcolor{Red}{marenamente}      & \textcolor{Green}{teneramente}   \\
S\textcolor{blue}{\textbf{2}}  & \textcolor{Red}{di}             & \textcolor{Green}{la}         & \textcolor{Green}{la}    \\
S\textcolor{blue}{\textbf{5}} & \textcolor{Red}{posa}        & \textcolor{Red}{posa} & \textcolor{Green}{sposa} \\
\midrule
\multicolumn{4}{c}{\textit{\underline{Soluzione:} AT\textcolor{Green}{tendere tenera}MEN\textcolor{Green}{te la} S\textcolor{Green}{posa}}} \\
\end{tabular}
\begin{tabular}{p{0.95\linewidth}}
\toprule
\toprule
\textit{\underline{Rebus:} B [Una figura geometrica]$^{~\text{D}1}$ [La si impugna per far girare un congegno]$^{~\text{D}2}$ DA [Le produce il rovo]$^{~\text{D}3}$} \\[2em]
\textit{\underline{Chiave di lettura:} \textcolor{blue}{\textbf{10 7 1' 5}}} \\[4pt]
\midrule
\end{tabular}
\begin{tabular}{lccc}
\textbf{Step} & \textbf{GPT-4o} & \textbf{Claude 3.5S} & \textbf{Phi-3} \\
D1 & \textcolor{Red}{cerchio}         & \textcolor{Red}{rombo}         & \textcolor{Green}{ellissi}     \\
D2 & \textcolor{Green}{manovella}   & \textcolor{Green}{manovella}   & \textcolor{Red}{leva}   \\
D3 & \textcolor{Green}{more}       & \textcolor{Green}{more}       & \textcolor{Green}{more}    \\
\midrule
S\textcolor{blue}{\textbf{10}}  & \textcolor{Red}{Bcerchiomanovella}     & \textcolor{Red}{Bromomanov} & \textcolor{Red}{Bellissile}     \\
S\textcolor{blue}{\textbf{7}}  &           & \textcolor{Red}{elladam}      & \textcolor{Red}{vadamore}   \\
S\textcolor{blue}{\textbf{1'}}  & \textcolor{Green}{d'}             & \textcolor{Red}{o'}         & \textcolor{Red}{'}    \\
S\textcolor{blue}{\textbf{5}} & \textcolor{Green}{amore}        & \textcolor{Red}{more} & \textcolor{Red}{remo} \\
\midrule
\multicolumn{4}{c}{\textit{\underline{Soluzione:} B\textcolor{Green}{ellissima novella} D' A\textcolor{Green}{more}}} \\
\bottomrule
\end{tabular}
\caption{Examples of LLM generations for rebuses by \textit{Baruffa, Rebus n. 12} (top), \textit{Contini C., La Settimana Enigmistica n. 4102} (mid) and \textit{Liosca, La Settimana Enigmistica n. 4581} (bottom). \textcolor{Green}{Correct guesses} and \textcolor{Red}{errors} and denoted for predicted first pass definitions (D$_{1,\dots,N}$) and solution words (S$_i$, with $i$ being the $i$-th solution key value).}
\vspace{-10pt}
\label{tab:more-ex-2}
\end{table}

\end{document}